\pdfoutput=1

\documentclass[11pt]{article}

\usepackage[preprint]{acl}

\usepackage{times}
\usepackage{latexsym}

\usepackage[T1]{fontenc}

\usepackage[utf8]{inputenc}

\usepackage{microtype}
\usepackage{amssymb}
\usepackage{amsmath}
\usepackage{amsthm}
\usepackage{booktabs}
\usepackage{enumitem}
\usepackage{color}
\usepackage{soul}
\usepackage{multirow}

\usepackage{inconsolata}

\usepackage{graphicx}
\usepackage[switch]{lineno}
\usepackage{tikz}
\usepackage{pgfplots}
\usepgfplotslibrary{polar}
\pgfplotsset{compat=1.18}

\title{MDToC: Metacognitive Dynamic Tree of Concepts for Boosting Mathematical Problem-Solving of Large Language Models}

\author{
  \textbf{Tung Duong Ta\textsuperscript{1}},
  \textbf{Tim Oates\textsuperscript{1}},
  \textbf{Thien Van Luong\textsuperscript{2}},
  \textbf{Huan Vu\textsuperscript{2}},
  \textbf{Tien Cuong Nguyen\textsuperscript{3}}
\\
\\
  \textsuperscript{1}University of Maryland, Baltimore County,
  \textsuperscript{2}National Economics University, Vietnam,
  \textsuperscript{3}VNPT AI, Vietnam
\\
  \texttt{\{dta1, oates\}@umbc.edu},
  \texttt{\{thienlv, huanv\}@neu.edu.vn},
  \texttt{nguyentiencuong@vnpt.vn}
}

\begin{document}
\maketitle
\begin{abstract}
Despite advances in mathematical reasoning capabilities, Large Language Models (LLMs) still struggle with calculation verification when using established prompting techniques. We present MDToC (Metacognitive Dynamic Tree of Concepts), a three-phase approach that constructs a concept tree, develops accuracy-verified calculations for each concept, and employs majority voting to evaluate competing solutions. Evaluations across CHAMP, MATH, and Game-of-24 benchmarks demonstrate our MDToC's effectiveness, with GPT-4-Turbo achieving 58.1\% on CHAMP, 86.6\% on MATH, and 85\% on Game-of-24 - outperforming GoT by 5\%, 5.4\%, and 4\% on all these tasks, respectively, without hand-engineered hints. MDToC consistently surpasses existing prompting methods across all backbone models, yielding improvements of up to 7.6\% over ToT and 6.2\% over GoT, establishing metacognitive calculation verification as a promising direction for enhanced mathematical reasoning.
\end{abstract}

\section{Introduction}

Large Language Models (LLMs) like GPT-4 \citep{achiam2023gpt} and Claude \citep{anthropic2024claude} demonstrate proficiency in various mathematical problems, excelling in easy to medium-difficulty tasks as evidenced by their performance on benchmarks such as GSM8k \citep{cobbe2021training} and SVAMP \citep{patel2021nlp}. However, their efficacy diminishes when faced with complex challenges presented in datasets such as MATH \citep{hendrycks2021measuring} and CHAMP \citep{mao2024champ}. In these demanding scenarios, LLMs often struggle with accurate multi-step reasoning and solution derivation. A key factor in this performance degradation is the models' propensity for errors in intermediate calculations and logical deductions \citep{patel2024multi, tyagi2024step}. These compounding inaccuracies result in poor performance on datasets featuring hard mathematical problems, opening a critical area for improvement in the multi-step reasoning capabilities of LLMs.

Researchers have widely adopted prompting techniques, particularly Chain-of-Thought (CoT) \citep{wei2022chain} and self-consistency CoT (SC-CoT) \citep{wang2023self},  to enhance LLMs' multi-step reasoning capabilities without additional training. These methods enable models to decompose complex reasoning processes into smaller steps, improving overall accuracy. In particular, CoT encourages articulation of thought processes, while SC-CoT generates multiple demonstrations with majority voting. However, these approaches have limitations:  CoT may constrain diverse problem-solving pathways, while SC-CoT lacks crucial evaluation of intermediate reasoning steps. This can lead to erroneous samples and inaccurate voting outcomes. As a result, using these prompting techniques, advanced LLMs such as GPT-4 suffer from poor performance. For example, GPT-4 with SC-CoT achieves only $9\%$ accuracy on the Game-of-24 task \citep{yao2024tree}. Therefore, it is essential to develop more robust reasoning methodologies.

\begin{figure}[t]
    \centering
    \includegraphics[width = .5\textwidth]{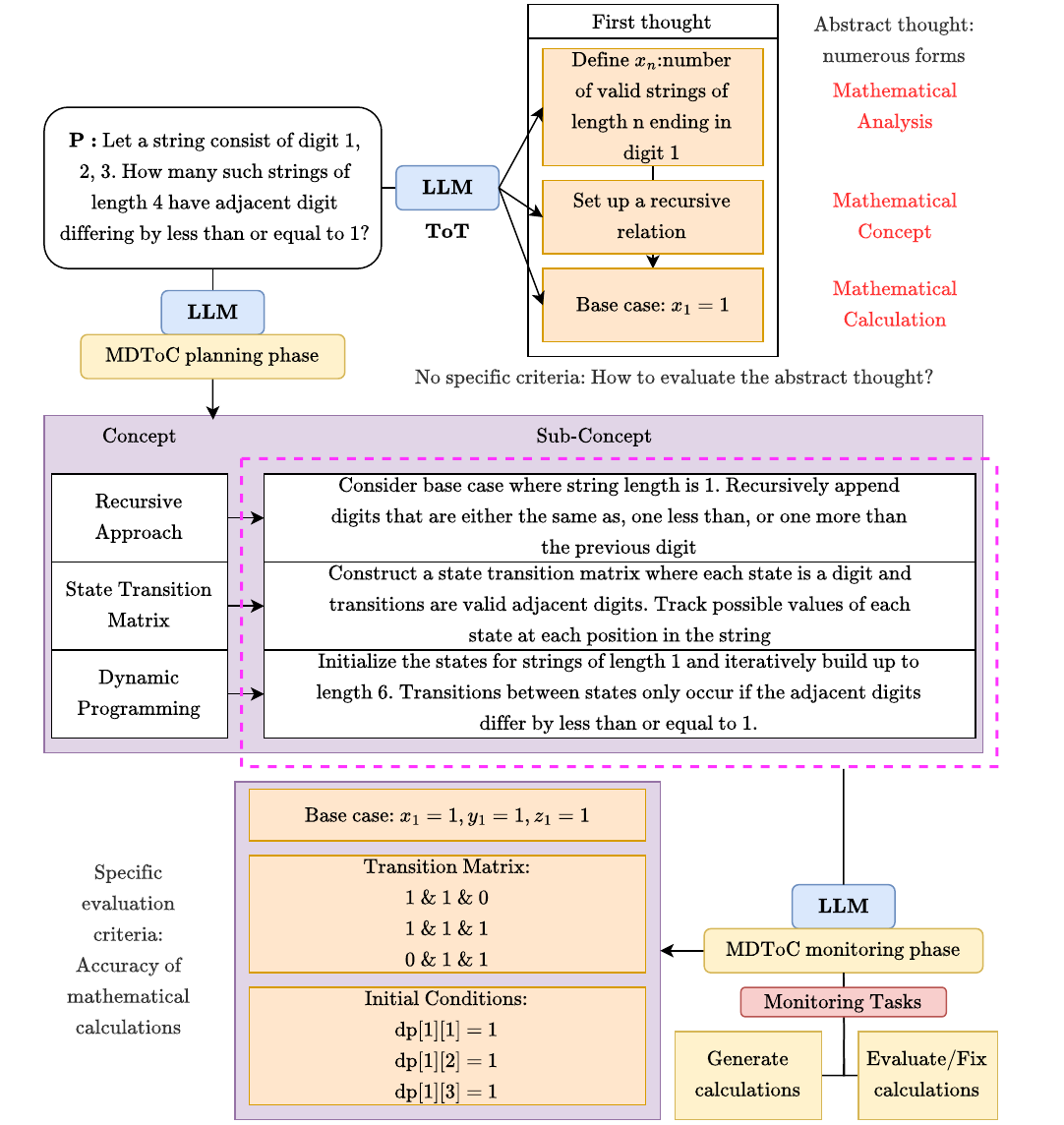}
    \caption{ToT prompting yields initial abstract thoughts (e.g., analyses, concepts, calculations; in \textcolor{red}{red}), which are challenging to evaluate due to the intangible nature of conceptual reasoning and the lack of specific criteria to measure their correctness or completeness. Our MDToC addresses these abstract thoughts by first generating concrete concepts and then producing relevant calculations for those concepts. We only evaluate the preciseness of the calculations through mathematical accuracy checks, enabling precise evaluation and thus improving problem-solving reliability.}
    \label{intro_exp}
\end{figure}

Recent hierarchical prompting techniques like Tree-of-Thoughts (ToT) \citep{yao2024tree} \citep{long2023large} and Graph-of-Thoughts (GoT) \citep{besta2024graph, yao2023beyond} have advanced reasoning capabilities through structured thought representation and intermediate evaluation, achieving impressive results on complex tasks (74\% accuracy on Game-of-24 \citep{yao2024tree} and 89\% on Sequence-Sorting-64-elements \citep{besta2024graph}, respectively). However, as shown in Figure \ref{intro_exp}, these approaches suffer from ill-defined evaluation criteria for diverse thought forms (mathematical analysis, concepts, calculations), leading to heavy reliance on powerful LLMs like GPT-4 that produce approximated and unreliable evaluation scores. This standardization challenge creates vulnerabilities in the critical processes of thought selection and connection pruning, while the need for domain-specific customization limits generalizability across different problem types.



Amidst the numerous successes of the CoT, ToT, and GoT prompting techniques, there have been several explorations of cognitive prompting methods for mathematical problem-solving \citep{fagbohun2024empirical}. In the field of psychology, metacognition enables individuals to reflect on and critically analyze their thought processes \citep{lai2011metacognition}. Recent research has enhanced model capabilities with metacognitive processes for natural language understanding tasks. For example,  \citep{wang2023metacognitive} demonstrated that LLMs prompted with metacognitive thinking outperformed previous techniques such as zero-shot \citep{kojima2022large} \citep{brown2020language} or CoT prompting \citep{wei2022chain} across various NLP tasks. \citep{zhou2024metacognitive} highlighted the effectiveness of the metacognitive approach in improving the retrieval-augmented generation process for LLMs. However, the application of metacognition to mathematical problem-solving remains relatively unexplored, with notable exceptions such as \citep{didolkar2024metacognitive}, who showed that metacognitive approaches enhance mathematical reasoning in LLMs by reflecting on clustered math skills and thereby providing relevant in-context examples. Our work extends this research direction by applying metacognition to LLMs for solving mathematical problems.

Motivated by the aforementioned background, we propose MDToC (Metacognitive Dynamic Tree of Concepts), a novel three-phase prompting technique that transforms abstract thoughts into concepts and evaluable calculations. MDToC employs a depth-two concept tree in the planning phase to explore diverse mathematical concepts while constraining the solution space, followed by a monitoring phase that expands sub-concepts with calculation steps using four specialized LLMs, and concludes with a reviewing phase utilizing majority voting following the self-consistency voting mechanism \cite{chen2023universal}. This comprehensive framework has demonstrated significant effectiveness, with GPT-3.5+MDToC achieving 39.5\% accuracy on CHAMP (outperforming GPT-3.5 with annotated concepts by 5.1\%) and GPT-4o-mini+MDToC attaining 75\% accuracy on Game-of-24 (surpassing GPT-4o-mini with ToT by 19\%).

\begin{figure*}[htbp]
    \centering
    \includegraphics[width = \textwidth]{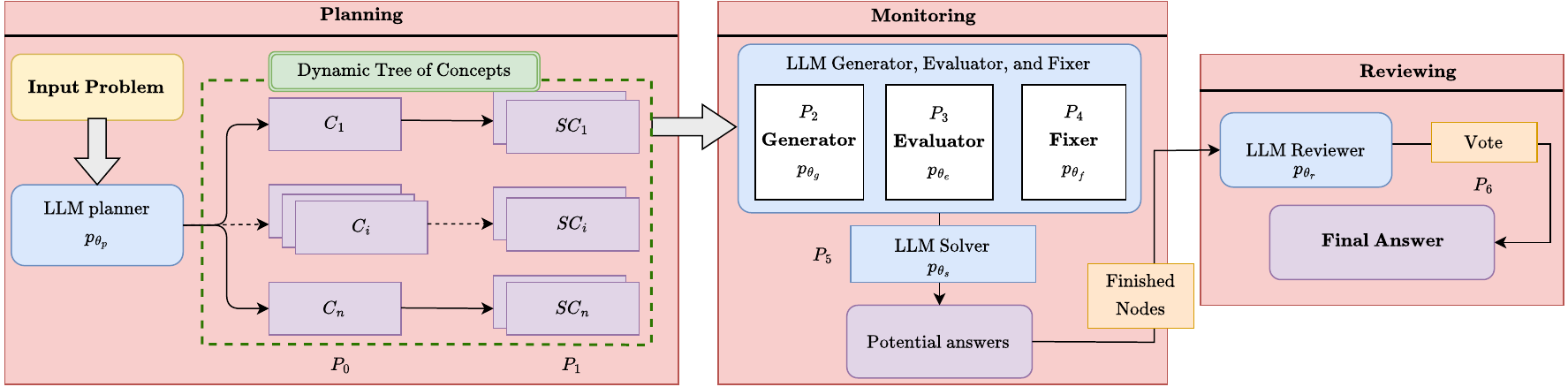}
    \caption{Proposed MDToC prompting structure. $C$ represents the first-depth concept, while $SC$ represents the second-depth sub-concept. $P_0$ and $P_1$ are prompts used in the planning phase shown in Figure \ref{p_plan}, while $P_2$, $P_{3}$, $P_4$, and $P_5$ are prompts used in the monitoring phase given in Figure \ref{p_mnt}. Prompts $P_6$ is the prompt in the review phase, as shown in Fig.~\ref{p_review}.}
    \label{method}
\end{figure*}

\section{Related Work}
\subsection{Prompting techniques}
Let $p_{\theta}$ be a LLM parameterized by $\theta$. Two of the most common prompting techniques for mathematical reasoning with $p_{\theta}$ are described below.

\begin{enumerate}
    \item \textbf{Chain-of-Thought}. 
    Given an input problem $x$, the LLM is guided through a sequence $c$ of reasoning steps to produce an answer $y$ \cite{wei2022chain}. Specifically, the sequence $c$ of reasoning steps is generated by the LLM based on the input problem, expressed as $c \sim p_{\theta}(c|x)$. The final answer $y$ is then generated by the LLM based on both the input problem and the sequence of reasoning steps expressed as $y \sim p_{\theta}(y|x, c)$. 

     \item \textbf{Tree-of-Thought}.
    Given an input problem $x$, the LLM navigates a tree structure where each node $i$ represents a state $s = \left[x, z_{1...i}\right]$, with $z_{1...i}$ denoting a sequence of thoughts along the current path \cite{yao2024tree}. The LLM generates a new thought $z_{i+1}$ based on the current state $s$, expressed as $z_{i+1} \sim p_{\theta}(z_{i+1} | x, z_{1...i})$. A new node with the state $s' = \left[x, z_{1...i+1}\right]$ is then appended to the current node $i$ in the tree.
    
    Each state $s$ in a set of states $S$ undergoes evaluation through either numerical values or voting to determine the viability of further path exploration. The numerical evaluation $V(p_{\theta}, s)$ is expressed as $V(p_{\theta}, s) \sim p_{\theta}(v | s) \text{ } \forall s \in S$, where $v$ is the numerical value. The voting evaluation $V(p_{\theta}, s)$ is expressed as $V(p_{\theta}, s) = 1
    \left[s = s^{\ast}\right] \text{ where } s^{\ast} \sim p^{vote}_{\theta}(s^{\ast} | S)$. In this context, The LLM votes for state $s^{\ast}$ given the set of states $S$, employing the indicator function $ 1\left[s = s^{\ast}\right]$ to determine whether a state $s$ corresponds to the voted state $s^{\ast}$.

    \item \textbf{Graph-of-Thought}
    Given an input problem $x$, the LLM navigates a directed graph structure $G = (V, E)$, where $V$ is the set of vertices representing thoughts, and $E \subseteq V \times V$ is the set of edges representing dependencies among thoughts \cite{besta2024graph}. Denote that $V^{+}$ and $E^{+}$ represent newly added vertices and edges, while $V^{-}$ and $E^{-}$ denote removed vertices and edges, respectively. Unless stated otherwise, $V^{-} = E^{-} = \emptyset$. The graph of thoughts is manipulated through three primary operations: aggregation, refinement, and generation.

    Following these operations, the graph is updated as $G' = T(G, p_{\theta}) = (V', E')$, where $V' = (V \cup V^{+}) \backslash V^{-}, \text{ } E' = (E \cup E^{+}) \backslash E^{-}$. Each node $v$ in graph $G$ is subsequently evaluated by the LLM using either a scoring or ranking method. The scoring function is expressed as $s = E(p_{\theta}, v, G) \sim p_{\theta}(s | v, G)$, where $s$ denotes the score value of node $v$. Conversely, the ranking function is expressed as $\lbrace v_1, v_2, ..., v_h \rbrace = R(p_{\theta}, h, G) \sim p_{\theta}(\lbrace v_1, v_2, ..., v_h \rbrace | G, h)$, where $h$ represents the number of top-ranking thoughts to be returned.
\end{enumerate}

\subsection{Metacognition}
Metacognition—the ability to reflect on and regulate one's thought processes—plays a crucial role in advanced problem-solving and decision-making. It serves as an overarching framework guiding the effective application of cognitive strategies. This study aims to endow language models with a simulated metacognitive process, mimicking the human capacity for ``thinking about thinking''. Our MDToC approach employs a hierarchical prompting structure incorporating three foundational stages of metacognition: planning, monitoring, and reviewing \cite{ku2010metacognitive}, specifically designed for mathematical problem-solving.

The planning stage creates a conceptual roadmap by establishing strategies and approaches. During monitoring, we implement a metacognitive mechanism that enables self-evaluation and correction of calculations in progress. The final reviewing stage examines solutions, filtering out empty results and identifying the most frequently occurring valid answer.

\section{Methodology}
This research introduces a novel prompting approach, called MDToC, utilizing a dynamic tree of concepts within a tripartite metacognitive framework of planning, monitoring, and reviewing. This method addresses limitations in existing hierarchical prompting techniques for LLMs, such as unreliable evaluations of abstract thoughts and lack of generalizability. Our approach enables the exploration of diverse reasoning paths and selects the final solution through majority voting. Figure \ref{method} shows a visual representation of our methodology.

\subsection{Planning}
During the initial planning phase, we construct a concept tree $T = (V, E)$ with a depth of two where $V$ is a set of nodes and $E$ is a set of edges. We begin by instructing our LLM planner $p_{\theta_p}$ with prompt $P_0$ to extract the objective of the question, called $q$, and generate $n$ distinct concepts $\lbrace c^{d=1}_1, c^{d=1}_2, ..., c^{d=1}_n \rbrace \sim p_{\theta_p} (q)$, where $d=1$ represents the first depth of the tree. Each $i$-th concept is articulated as a detailed sentence, providing a mathematical or programmatic response to $q$. We then incorporate $n$ concepts as nodes within the tree structure, where $V = \lbrace c^{d=1}_1, c^{d=1}_2, ..., c^{d=1}_n \rbrace$.

\begin{figure}[htbp]
    \centering
    \includegraphics[width = .45\textwidth]{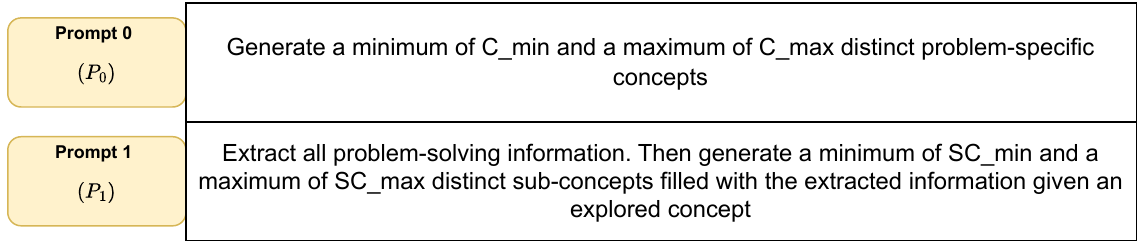}
    \caption{Prompts for the planning phase.}
    \label{p_plan}
\end{figure}

For each $i$-th concept at depth $d=1$, we further prompt our LLM planner $p_{\theta_p}$ with prompt $P_1$ to produce $m$ distinct sub-concepts $\lbrace c^{d=2}_{i1}, c^{d=2}_{i2}, ..., c^{d=2}_{im} \rbrace \sim p_{\theta_p} (q, c^{d=1}_i)$. These sub-concepts, each expressed in two detailed sentences, serve to elucidate and expand upon the $i$-th concept with question-solving information. Within our tree structure, we position each $j$-th sub-concept as a child node to its corresponding $i$-th concept. Figure \ref{p_plan} demonstrates our prompts for $P_0$ and $P_1$. Our tree $T$ is then expressed as:
\begin{align*}
    &V = \lbrace c^{d=1}_1, ..., c^{d=1}_n, c^{d=2}_{11}, ..., c^{d=2}_{1m}, ... , c^{d=2}_{n1}, ..., c^{d=2}_{nm} \rbrace \\
    & E = \lbrace (c^{d=1}_1, c^{d=2}_{11}), ..., (c^{d=1}_n, c^{d=2}_{nm}) \rbrace
\end{align*}.

\label{sec:planning}
The construction of our dynamic tree $T$ incorporates \textbf{four key parameters}: $C_{min}$, $C_{max}$, $SC_{min}$, and $SC_{max}$. These parameters define the concept range ($C_{min}$ and $C_{max}$) and sub-concept range ($SC_{min}$ and $SC_{max}$), as illustrated in prompts $P_0$ and $P_1$ of Figure \ref{p_plan}, respectively. The magnitude of these parameters directly correlates with the breadth of concept exploration, where larger values facilitate a more extensive conceptual landscape. Upon examining the annotated concepts in CHAMP \cite{mao2024champ} where each problem receives 3 hints, we decided to opt for a two-depth dynamic tree structure. We aim to prevent LLMs from excessively relying on generated concepts while fostering a more flexible problem-solving process that can adapt to various problem types and complexities.

\begin{figure}[htbp]
    \centering
    \includegraphics[width = .45\textwidth]{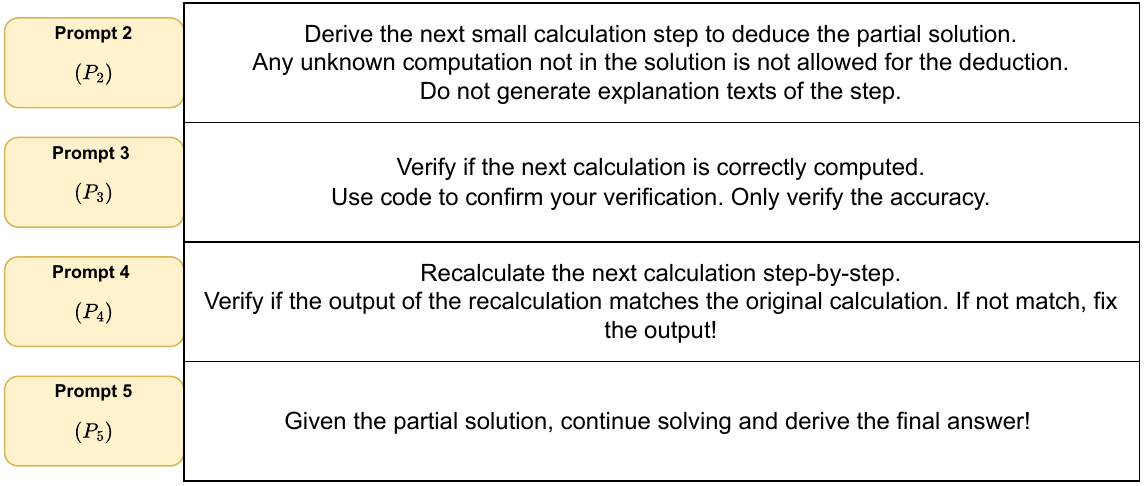}
    \caption{Prompts for the monitoring phase.}
    \label{p_mnt}
\end{figure}

\begin{figure*}[t]
    \centering
    \includegraphics[width=0.95\textwidth]{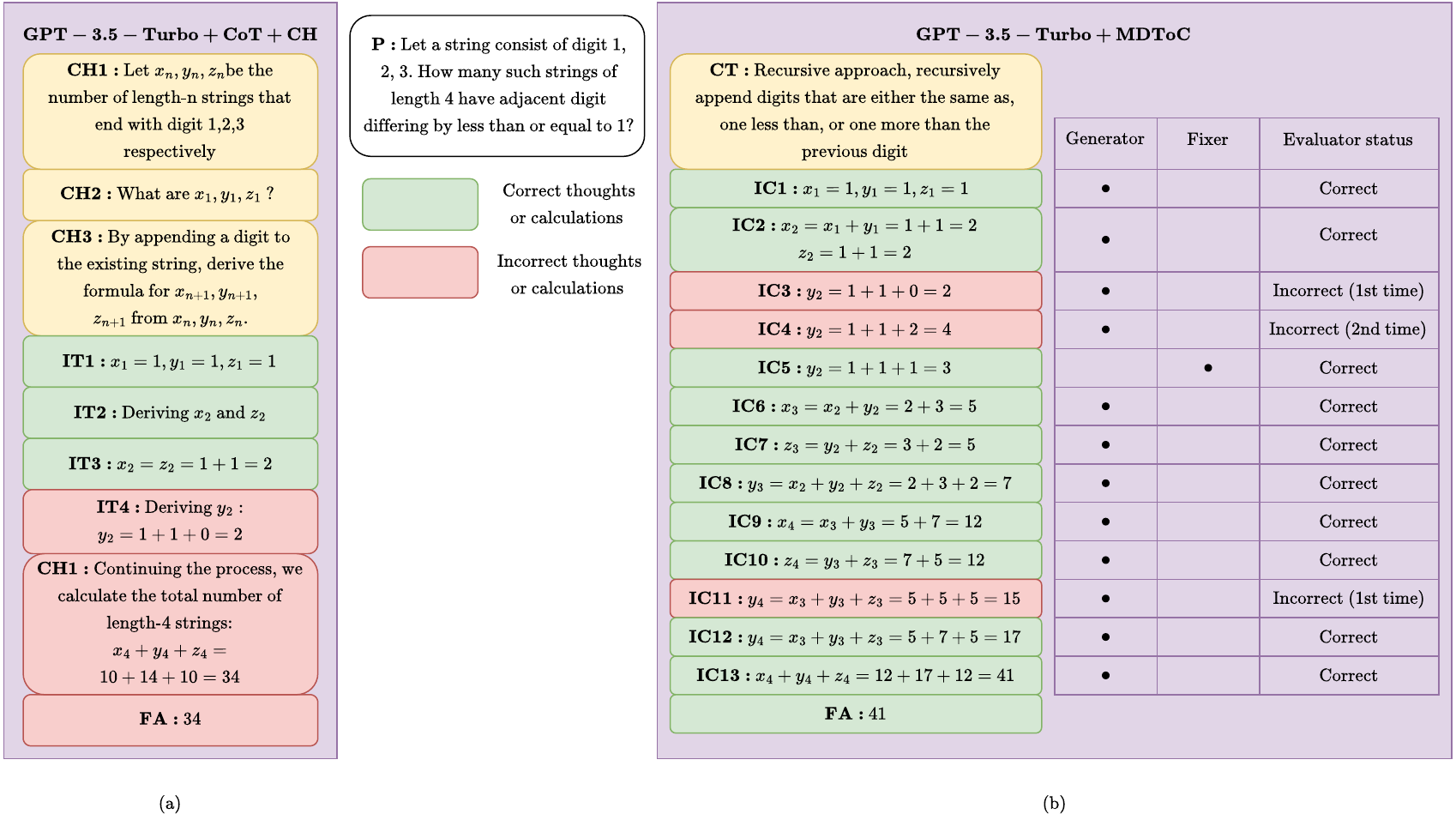}
    \caption{\textbf{Comparative analysis of reasoning steps: GPT-3.5-Turbo with CoT and CH versus GPT-3.5-Turbo with our MDToC approach}. Subfigure (a) displays GPT‑3.5-Turbo’s reasoning with CoT, supplemented by annotated concepts and hints \textbf{(CH)} intended to guide the model’s step-by-step reasoning; \textbf{IT} denotes intermediate thoughts, and \textbf{FA} indicates the final answer. Although these conceptual hints attempt to structure the problem-solving process, GPT‑3.5-Turbo still yields an incorrect count of \textbf{34}. Because there is no automatic mechanism to spot and correct mistakes in intermediate steps, the model’s calculation errors persist through to the final answer. Subfigure (b) shows our proposed concept tree (CT) approach under a multi-attempt evaluator–fixer framework, referred to here as MDToC. \textbf{IC} stands for intermediate calculations, and each is evaluated by an evaluator component. In this example, \textbf{IC3} and \textbf{IC4} are identified as incorrect, triggering the fixer to regenerate corrected values in \textbf{IC5}. This iterative refine-and-fix process avoids propagating calculation errors, ultimately yielding the correct final answer \textbf{FA} of \textbf{41}. Notably, this process requires no extra annotated hints —  only the concept tree plus repeated evaluation up to $2$ attempts, a threshold chosen to reduce the risk of model “hallucinations” (erroneous or fabricated steps).}
    \label{abl_study}
\end{figure*}

\subsection{Monitoring}
In the monitoring phase, we employ a ToT-based structure \cite{yao2024tree} to solve a concept tree in $t$ iterations. We denote a mathematical calculation as $\chi$. Unlike the traditional ToT approach which samples $k$ thoughts and selects a subset of them, our MDToC samples and evaluates all $k$ mathematical calculations with two LLM components: an LLM evaluator $p_{\theta_e}$ and an LLM generator $p_{\theta_{g}}$. The generator $p_{\theta_{g}}$ prompted with $P_2$ samples $k$ calculations, while the evaluator $p_{\theta_e}$ prompted with $P_{3}$ assesses the accuracy of each calculation. Specifically, when the evaluator $p_{\theta_e}$ identifies an error in a calculation, it returns a negative response ("No") accompanied by a detailed explanation of the error.

The $k$-th calculation $\chi^{d \geq 3}_{ijk}$ is sampled from the generator as $\chi^{d \geq 3}_{ijk} \sim p_{\theta_{g}} (\chi^{d \geq 3}_{ijk} | c^{d=1}_i, c^{d=2}_{ij}, \chi^{d - 1 \geq 3}_{ijk})$ where $\chi^{d - 1 \geq 3}_{ijk}$ represents previous calculations. The evaluation result $V(p_{\theta_e}, \chi^{d \geq 3}_{ijk})$ for the $k$-th calculation is expressed as $V(p_{\theta_e}, \chi^{d \geq 3}_{ijk}) \sim p_{\theta_e} (v_e, r_e | (c^{d=1}_i, c^{d=2}_{ij}, \chi^{d - 1 \geq 3}_{ijk}, \chi^{d \geq 3}_{ijk})]$, where $v_e$ is the binary result of $0$ or $1$ and $r_e$ is the evaluation reason when $v_e = 0$. If $v_e = 0$, the generator regenerates the calculation $\chi^{d \geq 3}_{ijk} \sim p_{\theta_{g}} (\chi^{d \geq 3}_{ijk} | c^{d=1}_i, c^{d=2}_{ij}, \chi^{d - 1 \geq 3}_{ijk}, \chi^{d \geq 3}_{ijk}, r_e)$.

We further introduce an LLM fixer $p_{\theta_f}$ prompted with $P_5$ to fix the errors in the $k$-th calculation as $ \chi^{d \geq 3}_{ijk}  \sim p_{\theta_f}(\chi^{d \geq 3}_{ijk} | \chi^{d \geq 3}_{ijk})$. Specifically, after the complete cycle of generation and evaluation of calculations, $p_{\theta_f}$  addresses and corrects any remaining calculation errors. Subsequently, these $k$ calculations are appended as nodes to the tree. After some iterations, we treat the current series of calculations as a partial solution and employ an LLM solver $p_{\theta_s}$ to resolve the partial solution. The solver's response is subsequently appended to the tree and marked as a 'Finished Node', thereby terminating the exploratory process. Figure \ref{p_mnt} illustrates our prompts $P_2$, $P_{3}$, $P_4$, and $P_5$.

\label{sec:monitoring}
To monitor our concept tree with calculations, we introduced \textbf{two additional parameters}: $c_s$, $t$. $c_s$ specifies the number of intermediate calculations generated, respectively, while $t$ represents the number of iterations. These parameters allow us to control the exploration behavior, balancing between breadth and depth.

\subsubsection{Example Analysis}
Figure \ref{abl_study} compares intermediate reasoning steps between GPT-3.5-Turbo with CoT + CH and GPT-3.5-Turbo with MDToC. It highlights challenges in recursive calculations requiring precise intermediate computations. Figure \ref{abl_study}a shows GPT-3.5-Turbo's performance is limited to accurate calculation of the base case only, with errors emerging in $y_2$ and cascading through subsequent steps. This leads to an incorrect final sum of $34$. 

Conversely, Figure \ref{abl_study}b illustrates the efficacy of our MDToC approach in mitigating computational errors. When applying MDToC, the likelihood of erroneous intermediate calculations is significantly reduced with $2$ evaluation and regeneration attempts. This is evidenced by the correct computation of $y_2$, accurately determined as the sum of $x_1 = 1$, $y_1 = 1$, and $z_1 = 1$, yielding the correct result of $3$. This precise evaluation of $x_2$, $y_2$, and $z_2$ serves as a crucial foundation, enabling accurate subsequent calculations for strings of lengths of $3$ and $4$, ultimately leading to the correct final answer of $41$.

\subsection{Reviewing}
\begin{figure}[htbp]
    \centering
    \includegraphics[width = .45\textwidth]{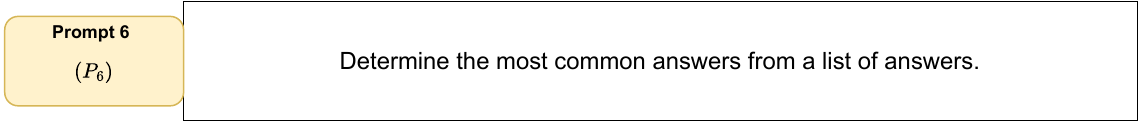}
    \caption{Prompts for the reviewing phase.}
    \label{p_review}
\end{figure}

In the last phase, we obtain the results from the monitored tree and select the top-voted answers. Given a list of solutions marked by 'Finished Node' as $A$, we use an LLM reviewer $p_{\theta_r}$ prompted with $P_6$ shown in Figure \ref{p_review} to conduct a majority voting of $A$. This stage involves finding the most common answer from $A$, returning the most common one as $\tilde{a} \sim p_{\theta_e}(A)$, where $\tilde{a}$ is the final solution to the objective $q$.

\section{Experiments}
\subsection{Dataset}
Our MDToC approach is evaluated using three datasets: CHAMP \cite{mao2024champ} (270 high school-level competition math problems across five categories, providing insight into concept tree-driven reasoning), MATH \cite{hendrycks2021measuring} (12,500 competition-level problems from sources like AIME \cite{aopsAIME} and AMC 10/12 \cite{aopsAMC10} covering various advanced topics including algebra, geometry, and number theory), and a subset of 100 challenging Game-of-24 puzzles (indexed 901-1000, selected for direct comparison with ToT approaches) \cite{yao2024tree}.

\subsection{Parameters}
\textbf{Planning phase} For the CHAMP and MATH datasets, we set $C_{min} = 2$, $C_{max} = 3$, $SC_{min} = 1$, and $SC_{max} = 2$ (see our \ref{sec:planning}). In contrast, all parameters were set to 1 for Game-of-24 experiments, as these problems require diverse calculation combinations rather than varied mathematical concepts.

\textbf{Monitoring phase} For the CHAMP and MATH datasets, we use $c_s = 2$ and $t = 10$ (see our \ref{sec:monitoring}), emphasizing in-depth concept decomposition and analysis. Game-of-24 employs $c_s = 10$ and $t = 4$, focusing on broader analysis of calculation combinations.

\subsection{Model Configuration}
In our experiments, we use OpenAI LLMs throughout. GPT-4o handles the planning phase for concept diversification and the review phase for response standardization and voting. For the monitoring phase's fixer component $p_{\theta_f}$ (see Fig.~\ref{method}), we deploy GPT-4o-mini for cost efficiency while maintaining accuracy. The generator, evaluator, and solver components (see Fig.~\ref{method}) utilize four different models (GPT-3.5-Turbo, GPT-4o-mini, GPT-4-Turbo, and GPT-4o) to enable direct comparison with other prompting methods. Therefore, GPT-3.5-Turbo+MDToC, GPT-4o-mini+MDToC, GPT-4-Turbo+MDToC, and GPT-4o+MDToC schemes each use their namesake model for monitoring phase components, while all sharing GPT-4o for planning/reviewing and GPT-4o-mini for fixing.

To demonstrate the fairness of these comparisons despite including GPT-4o and GPT-4o-mini, we analyzed token consumption across all models and phases for three datasets in Table~\ref{tab:tokens_usage}. The analysis shows that GPT-4o for the planning and reviewing phase uses less than 1\% of total tokens for both CHAMP and MATH datasets and approximately 2\% for Game-of-24. In the fixer component, GPT-4o-mini consumes 23,428, 21,599, and 1,945 tokens for CHAMP, MATH, and Game-of-24 respectively, accounting for about 6\% of total tokens in the monitoring phase. Since the remaining three GPT models are responsible for roughly 93\% of overall token usage, we can confidently make valid comparisons with alternative prompting techniques that utilize these four primary GPT models.

\begin{table}[h]
\caption{\label{tab:tokens_usage}Average tokens used per response of GPT-4o, GPT-3.5-Turbo, GPT-4o-mini, and GPT-4-Turbo across the planning, reviewing, and monitoring phases for the CHAMP, MATH, and Game-of-24 (G24).}
\begin{center}
\begin{tabular}{|c|p{1.5cm}|p{1.2cm}|p{1.1cm}|p{0.9cm}|}
\hline
\multirow{2}{*}{\textbf{GPT}} & \multirow{2}{*}{\textbf{Phase}} & \multicolumn{3}{c|}{\textbf{Dataset}}  \\
\cline{3-5}
& & CHAMP & MATH & G24 \\
\hline
\multirow{2}{*}{4o} & Planning & 2,671 & 2,292 & 655 \\
\cline{2-5}
& Reviewing & 346 & 424 & 127 \\
\hline
3.5-Turbo & \multirow{4}{1.5cm}{\centering Generator Evaluator Solver} & 548,202 & 472,175 & 36,151 \\
\cline{1-1}
\cline{3-5}
4o-mini & & 465,420 & 443,532 & 34,437\\
\cline{1-1}
\cline{3-5}
4-Turbo & & 433,588 & 378,745 & 33,804 \\
\cline{1-1}
\cline{3-5}
4o & & 367,091 & 316,275 & 29,668 \\
\hline
4o-mini & Fixer & 23,428 & 21,599 & 1,945 \\
\hline
\end{tabular}
\label{model_config}
\end{center}
\end{table}

\subsection{CHAMP evaluation}
Table~\ref{acc_eval} compares various prompting techniques across GPT-3.5-Turbo, GPT-4o-mini, GPT-4-Turbo, and GPT-4o models. While traditional approaches like zero-shot, CoT, five-shot prompting show modest results, incorporating concepts and hints into prompting techniques demonstrates greater success. CoT + CH improves accuracy, with partial solution provision (1/3 sln) offering further gains. For instance, GPT-4-Turbo achieves 53.0\% accuracy with CoT+CH, significantly outperforming its 37.8\% CoT and 43.1\% 5-shot results.
\begin{table}[htbp]
\caption{Comparative performance of different prompting approaches for various GPTs on the CHAMP dataset. '0-shot' and '5-shot' denote in-context examples in prompts; '1/3 sln' indicate the proportion of complete solution provided; CoT and CH represent Chain-Of-Thought and Annotated concepts and hints in prompts.}
\begin{center}
\begin{tabular}{|p{1.3cm}|c|c|c|c|}
\hline
\multirow{2}{*}{\textbf{Prompt}} & \multicolumn{4}{c|}{\textbf{GPT}} \\
\cline{2-5}
& 3.5-Turbo & 4o-mini & 4-Turbo & 4o\\
\hline
0-shot & 28.5 & 36.2 & 41.9 & 55.3\\
\hline
CoT & 29.6 & 36.5 & 37.8 & 55.2\\
\hline
5-shot & 34.8 & 38.7 & 43.1 & 56.5\\
\hline
CoT + CH & 34.4 & 42.3 & 53.0 & 60.0\\
\hline
1/3 sln & 33.7 & 43.4 & 53.7 & 63.5\\
\hline
ToT & 31.7 & 37.8 & 52.7 & 61.3\\
\hline
GoT & 30.5 & 39.6 & 53.1 & 60.9\\
\hline
Our MDToC & \textbf{39.5} & \textbf{48.3} & \textbf{58.1} & \textbf{68.2} \\
\hline
\end{tabular}
\label{acc_eval}
\end{center}
\end{table}

Our MDToC demonstrates superior performance across all models, achieving 39.5\% (GPT-3.5-Turbo), 48.3\% (GPT-4o-mini), 58.1\% (GPT-4-Turbo), and 65.2\% (GPT-4o). These results represent substantial improvements over ToT and GoT, with gains ranging from 5.4\% to 10.5\%, underscoring the effectiveness of dynamically structuring relevant concepts and employing a metacognitive feedback mechanism to refine calculation steps.

\subsection{MATH evaluation}
In Table \ref{math_eval}, 0-shot prompts yield 45.7\% for GPT-3.5-Turbo, 71.4\% for GPT-4o-mini, 72.6\% for GPT-4-Turbo, and 81.5\% for GPT-4o on the MATH dataset. Adding reasoning steps through CoT improves these scores slightly (e.g., from 45.7\% to 48.6\% for GPT-3.5-Turbo), while increasing the number of examples (5-shot) provides further gains (up to 79.3\% for GPT-4-Turbo and 87.1\% for GPT-4o). The ToT and GoT method surpass standard few-shot prompts for the two more advanced models, pushing GPT-4-Turbo to 80.4\% and 81.2\% and GPT-4o to 87.1\% and 87.6\% — a notable jump of around 7\% from CoT.

Even so, our MDToC outperforms all these strategies, achieving 60.8\% for GPT-3.5-Turbo, 83.8\% for GPT-4o-mini, 86.6\% for GPT-4-Turbo, and 89.5\% for GPT-4o. Compared to ToT, our MDToC provides an extra 7.6-\% boost for GPT-3.5-Turbo, 5.3\% for GPT-4o-mini, and 6.2\% for GPT-4-Turbo. These results strengthen our claim that our MDToC not only overcomes the evaluation constraints in ToT but also achieves more robust performance than purely tree-based approaches like ToT.
\begin{table}[hbtp]
\caption{Comparative performance of different prompting approaches for various GPTs on the MATH dataset}
\begin{center}
\begin{tabular}{|p{1.3cm}|c|c|c|c|}
\hline
\multirow{2}{*}{\textbf{Prompt}} & \multicolumn{4}{c|}{\textbf{GPT}} \\
\cline{2-5}
& 3.5-Turbo & 4o-mini & 4-Turbo & 4o \\
\hline
0-shot & 45.7 & 71.4 & 72.6 & 81.5\\
\hline
CoT & 48.6 & 72.4 & 73.3 & 81.6\\
\hline
5-shot & 54.3 & 77.1 & 79.5 & 82.4\\
\hline
ToT & 53.2 & 78.5 & 80.4 & 87.1\\
\hline
GoT & 51.8 & 80.0 & 81.2 & 87.6\\
\hline
Our MDToC & \textbf{60.8} & \textbf{83.8} & \textbf{86.6} & \textbf{89.5} \\
\hline
\end{tabular}
\label{math_eval}
\end{center}
\end{table}

\subsection{Game-of-24 evaluation}

Table \ref{game24-acc} now reports six prompting strategies—0‑shot, CoT, 5‑shot, ToT, GoT, and our MDToC—evaluated on four models (GPT‑3.5‑Turbo, GPT‑4o‑mini, GPT‑4‑Turbo, and the new GPT‑4o). The broad pattern remains: minimal prompting (0‑shot or CoT) yields very low accuracy (2–10\%), while adding a handful of demonstrations (5‑shot) produces a modest gain (6–18\%). ToT then brings a significant jump for GPT‑4o‑mini to 56\%, GPT‑4‑Turbo to 74\%, and GPT‑4o to 88\%. GoT raises scores further to 62\%, 81\%, and 90\% for GPT‑4o‑mini, GPT‑4‑Turbo, and GPT‑4o, respectively; even GPT‑3.5‑Turbo climbs from 19\% (ToT) to 21\% (GoT).

Despite these gains, our MDToC still delivers the best accuracy across the board: 30\% (+9\% over GoT) on GPT‑3.5‑Turbo, 75\% (+13\%) on GPT‑4o‑mini, 85\% (+4\%) on GPT‑4‑Turbo, and a top‑tied 90\% on GPT‑4o. These margins underscore the critical role of MDToC’s evaluator $p_{\theta_e}$ and fixer $p_{\theta_f}$, which detect and repair erroneous intermediate expressions involving the four numbers in Game‑of‑24, maintaining logical consistency and reducing hallucinations throughout the reasoning chain.

\begin{table}[h]
\caption{Comparative performance of different prompting approaches for various GPTs on the Game-of-24 dataset.}
\begin{center}
\begin{tabular}{|p{1.3cm}|c|c|c|c|}
\hline
\multirow{2}{*}{\textbf{Prompt}} & \multicolumn{4}{c|}{\textbf{GPT}} \\
\cline{2-5}
& 3.5-Turbo & 4o-mini & 4-Turbo & 4o \\
\hline
0-shot & 2 & 3 & 4 & 10\\
\hline
CoT & 3 & 3 & 4 & 9\\
\hline
5-shot & 6 & 8 & 10 & 18\\
\hline
ToT & 19 & 56 & 74 & 88\\
\hline
GoT & 21 & 62 & 81 & 90\\
\hline
Our MDToC & \textbf{30} & \textbf{75} & \textbf{85} & \textbf{90}\\
\hline
\end{tabular}
\label{game24-acc}
\end{center}
\end{table}

\section{Discussion}
\subsection{Cost analysis of our MDToC configuration}

\begin{table}[htbp]
    \centering
    \begin{tabular}{|p{1.3cm}|c|p{0.7cm}|p{0.7cm}|p{0.6cm}|p{0.6cm}|}
    \hline
       \multirow{2}{*}{\textbf{Dataset}} & \multirow{2}{*}{\textbf{GPT}} & \multicolumn{2}{c|}{\textbf{Cost (\$)}} & \multicolumn{2}{c|}{\textbf{Accuracy}}  \\
       \cline{3-6}
         & & w & w/o & w & w/o \\
        \hline
        \multirow{4}{1.3cm}{\centering \textbf{MATH}} & 3.5-Turbo & 0.58 & 0.62 & \textbf{60.8} & 54.1 \\
        \cline{2-6}
        & 4-Turbo & 19.79 & 21.16 & 86.6 & \textbf{86.9} \\
        \cline{2-6}
        & 4o-mini & 0.23 & 0.23 & \textbf{83.8} & 82.9 \\
        \cline{2-6}
        & 4o & 3.5 & 3.7 & 89.5 & \textbf{89.7} \\
        \hline
        \hline
        \multirow{4}{1.3cm}{\centering \textbf{CHAMP}} & 3.5-Turbo & 0.69 & 0.72 & \textbf{39.5} & 33.9\\
        \cline{2-6}
        & 4-Turbo & 22.59 & 24.15 & 58.1 & \textbf{59.0} \\
        \cline{2-6}
        & 4o-mini & 0.24 & 0.24 & \textbf{48.3} & 47.1\\
        \cline{2-6}
        & 4o & 3.94 & 4.08 & 68.2 & \textbf{68.5}\\
        \hline
        \hline
        \multirow{4}{1.3cm}{\centering \textbf{G24}} & 3.5-Turbo & 0.03 & 0.05 & \textbf{30} & 25\\
        \cline{2-6}
        & 4-Turbo & 1.79 & 1.92 & \textbf{85} & 85\\
        \cline{2-6}
        & 4o-mini & 0.02 & 0.02 & \textbf{75} & 74\\
        \cline{2-6}
        & 4o & 0.37 & 0.39 & \textbf{90} & 90\\
        \hline
    \end{tabular}
    \caption{Per‑case cost and accuracy for our MDToC with GPT models on MATH, CHAMP, and Game-of-24 (G24). W stands for using GPT-4o-mini in the fixer component and GPT-4o in the planning and reviewing phase in our MDToC configuration. W/o stands for using the same LLMs across all components.}
    \label{cost_analysis}
\end{table}
Replacing the planner and reviewer with GPT-4o and the fixer with GPT-4o-mini reduces costs while maintaining or improving accuracy across all backbone models and datasets tested (shown in Table \ref{cost_analysis}). For GPT-3.5-Turbo, costs decreased while accuracy increased on MATH (54.1\% to 60.8\%), CHAMP (33.9\% to 39.5\%), and Game-of-24 (25\% to 30\%). GPT-4-Turbo saw cost reductions with minimal accuracy changes on all datasets, while GPT-4o as backbone showed 5-7\% cost savings with negligible accuracy differences ($\leq$0.3\%). These results demonstrate that delegating auxiliary roles to lighter models is an effective strategy for reducing computational expenses without compromising performance, with weaker models particularly benefiting from the diverse planning concepts provided by GPT-4o.

\subsection{Hyperparameter Sensitivity Test}
\begin{table}[h]
    \centering
    \begin{tabular}{|p{1.1cm}|p{1.0cm}|p{1.0cm}|p{0.7cm}|p{0.9cm}|p{0.7cm}|}
    \hline
    ($C_{min}$, $C_{max}$, $SC_{min}$, $SC_{max}$) & ($c_s$, $t$) & MATH Acc & G24 Acc & MATH Cost & G24 Cost \\
    \hline
    (2,4,1,2) & (2,10) & \textbf{89.5\%} & 87\% & \textbf{\$3.5} & \$0.7 \\
    \hline
    (3,5,1,2) & (2, 10) & 89.6\% & 88\% & \$4.9 & \$0.8 \\
    \hline
    (3,5,2,4) & (3, 15) & 89.9\% & 88\% & \$5.8 & \$1.0 \\
    \hline
    (1,1,1,1) & (15, 5) & 83.4\% & \textbf{90\%} & \$2.6 & \textbf{\$0.4} \\
    \hline
    \end{tabular}
    \caption{Hyperparameters (see our \ref{sec:planning} and \ref{sec:monitoring}) in MDToC with GPT-4o on MATH and Game-of-24 (G24). Acc stands for Accuracy.}
    \label{hyperparameters}
\end{table}
Table \ref{hyperparameters} shows important hyperparameter combinations. Increasing concept exploration (up to 5 concepts ($C_{\text{max}} = 5$ and 20 subconcepts in total ($SC_{\text{max}} = 4$) marginally improves the accuracy of our MDToC on MATH (0.4\%), confirming our finding that about 4 concepts ($C_{\text{max}} = 4$) subconcepts are sufficient while reducing the cost significantly (from \$5.8 to \$3.5). For Game-of-24, the accuracy slightly decreases (90\% to 88\%) with less broad explorations, indicating this dataset does not  benefit from concept variety. Meanwhile, despite of cost decrease (down to \$2.6), limited concepts with broad computational exploration reduce MATH accuracy (to 83.4\%). However, these suit exploration-based datasets like Game-of-24.

\section{Conclusion}
Our proposed MDToC framework enhances mathematical reasoning in LLMs through structured metacognition—planning, monitoring, and reviewing. It outperforms ToT and GoT techniques, achieving up to 11\% higher accuracy on Game-of-24 and showing consistent improvements on CHAMP and MATH. Our MDToC excels in calculation-intensive tasks through dynamic concept structuring and iterative error correction, establishing a foundation for future research in complex problem-solving.

\section{Limitations}
Despite MDToC's superior performance compared to previous prompting methodologies, several notable limitations affect its practical implementation. First, our metacognitive calculation approach exhibits domain-specific constraints, particularly in mathematical fields such as geometry, where spatial reasoning predominates over calculation verification. In such domains, the iterative verification processes central to MDToC may offer diminishing returns, as the primary cognitive challenges relate to geometry visualization rather than computational validation.

Second, the performance improvements delivered by our MDToC incur computational costs. As demonstrated in our analyses (see Tables \ref{tab:tokens_usage} and \ref{cost_analysis}), the method introduces significant resource overhead, particularly when implementing MDToC with expensive LLMs such as GPT-4-Turbo. Token consumption for CHAMP and MATH benchmark problems reaches approximately $450,000$ tokens per problem, translating to approximately $\$20$ per problem—a cost scale that may prove prohibitive for educational institutions and research organizations with limited budgets. These economic constraints potentially restrict MDToC's deployment in resource-limited environments.

\bibliography{custom}

\appendix

\section{Appendix}
\subsection{Complexity Analysis}
\label{sec:complexity}
Table \ref{Complexity_Analysis} demonstrates that our MDToC achieves higher accuracy than ToT. On the MATH benchmark with GPT-4-Turbo, our MDToC with GPT-4-Turbo raises accuracy from 80.4\% to 86.6\% (+6.2 \% gain) while consuming only 16\% more tokens (346k $\rightarrow$ 403k) and about 40 seconds of extra time — yielding roughly 1\% of extra accuracy for every 2.5\% of extra tokens. A similar pattern appears on CHAMP: our MDToC with GPT-4-Turbo delivers 5.4\% gain (52.7\% $\rightarrow$ 58.1\%) for just 14\% more tokens and a 0.2-minute latency increase. On the Game-of-24 task, our MDToC converts a five-fold token increase into an 11\% accuracy jump (74\% $\rightarrow$ 85\%) for GPT-4-Turbo while keeping runtime under 2 minutes. With GPT‑4o, for example, our MDToC converts a 28\% rise in tokens on MATH into 2.4\% accuracy gain over ToT (87.1\% $\rightarrow$ 89.5\%) and turns a 26\% token increase on CHAMP into 6.9\% accuracy gain (61.3\% $\rightarrow$ 68.2\%). These results confirm that the added metacognitive steps for calculation evaluation pay for themselves: each additional cost generates more correct answers than ToT can perform with the same model.
\begin{table}[h]
    \centering
    \begin{tabular}{|p{1.0cm}|c|c|c|p{0.7cm}|p{0.7cm}|p{0.6cm}|}
        \hline
        \textbf{Dataset} & \textbf{Prompt} & \textbf{GPT} & \textbf{Token} & \textbf{Cost} & \textbf{Time} & \textbf{Acc} \\
    \hline
    \multirow{4}{0.9cm}{\centering \textbf{MATH}} & \multirow{2}{*}{MDToC} & 4-Turbo & 403,060 & 19.79 & 12.1 & \textbf{86.6} \\
    \cline{3-7}
    & & 4o & 286,935 & 3.5 & 11.1 & \textbf{89.5} \\
    \cline{2-7}
     & \multirow{2}{0.9cm}{ToT} & 4-Turbo & \textbf{346,193} & \textbf{12.98} & \textbf{11.5} & 80.4 \\
    \cline{3-7}
     & & 4o & \textbf{223,088} & \textbf{2.79} & \textbf{11.0} & 87.1 \\
    \hline
    \hline
    \multirow{4}{0.9cm}{\centering \textbf{CHA-MP}} & \multirow{2}{*}{MDToC} & 4-Turbo & 460,033 & 22.59 & 12.3 & \textbf{58.1}\\
    \cline{3-7}
    & & 4o & 315,561 & 3.94 & 11.4 & \textbf{68.2} \\
    \cline{2-7}
     & \multirow{2}{*}{ToT} & 4-Turbo & \textbf{404,881} & \textbf{15.18} & \textbf{12.1} & 52.7 \\
    \cline{3-7}
    & & 4o & \textbf{249,713} & \textbf{3.12} & \textbf{11.1} & 61.3 \\
    \hline
    \hline
    \multirow{4}{*}{\centering \textbf{G24}} & \multirow{2}{*}{MDToC} & 4-Turbo & 36,531 & 1.79 & 1.8 & \textbf{85} \\
    \cline{3-7}
    & & 4o & 29,464 & 0.37 & 1.7 & \textbf{90} \\
    \cline{2-7}
     & \multirow{2}{*}{ToT} & 4-Turbo & \textbf{6,958} & \textbf{0.74} & \textbf{1.0} & 74 \\
    \cline{3-7}
    & & 4o & \textbf{6,179} & \textbf{0.08} & \textbf{1.0} & 88 \\
    \hline
\end{tabular}
\caption{Token and cost analysis of our MDToC and ToT on MATH, CHAMP, and Game-of-24 (G24). The token and cost are \textbf{per case}. Time is measured in \textbf{minutes}. Acc stands for Accuracy.}
\label{Complexity_Analysis}
\end{table}

\subsection{LLM Benchmark}
\begin{table}[htbp]
    \centering
    \begin{tabular}{|p{1.0cm}|p{1.1cm}|c|c|c|}
        \hline
        \multirow{2}{*}{\textbf{LLM}} & \multirow{2}{*}{\centering \textbf{Prompt}} & \multicolumn{3}{c|}{\textbf{Dataset}} \\
        \cline{3-5}
        & & CHAMP & MATH & G24 \\
        \hline
        \multirow{3}{0.9cm}{\centering GPT-3.5-Turbo} & ToT &  31.7 & 51.2 & 19 \\
        \cline{2-5}
        & GoT &  32.3 & 51.6 & 21 \\
        \cline{2-5}
        & MDToC & \textbf{33.9} & \textbf{54.1} & \textbf{25} \\
        \hline
        \hline
        \multirow{3}{0.9cm}{\centering GPT-4-Turbo} & ToT & 52.7 & 80.4 & 79 \\
        \cline{2-5}
        & GoT & 54.2 & 80.9 & 81 \\
        \cline{2-5}
        & MDToC & \textbf{59.0} & \textbf{86.9} & \textbf{85} \\
        \hline
        \hline
        \multirow{3}{0.9cm}{\centering GPT-4o-mini} & ToT & 37.8 & 78.5 & 56 \\
        \cline{2-5}
        & GoT & 39.3 & 79.0 & 62 \\
        \cline{2-5}
        & MDToC & \textbf{47.1} & \textbf{82.9} & \textbf{74} \\
        \hline
        \hline
        \multirow{3}{0.9cm}{\centering GPT-4o} & ToT & 61.3 & 87.1 & 88 \\
        \cline{2-5}
        & GoT & 63.4 & 87.9 & 90 \\
        \cline{2-5}
        & MDToC & \textbf{68.5} &  \textbf{89.7} & \textbf{90} \\
        \hline
        \hline
        \multirow{3}{0.9cm}{\centering Mistral-7B} & ToT & 18.1 & 23.5 & 8 \\
        \cline{2-5}
        & GoT & 18.8 & 24.2 & 9 \\
        \cline{2-5}
        & MDToC & \textbf{19.7} & \textbf{27.2} & \textbf{9}\\
        \hline
        \hline
        \multirow{3}{0.9cm}{\centering Mistral 8x22B} & ToT & 31.9 & 52.7 & 22 \\
        \cline{2-5}
        & GoT & 32.4 & 53.2 & 23 \\
        \cline{2-5}
        & MDToC & \textbf{34.5} & \textbf{55.4} & \textbf{26} \\
        \hline
        \hline
        \multirow{3}{0.9cm}{\centering Llama-3 8B} & ToT & 19.2 & 26.8 & 7  \\
        \cline{2-5}
        & GoT & 20.9 & 27.5 & 8 \\
        \cline{2-5}
        & MDToC & \textbf{21.5} & \textbf{30.6} & \textbf{10}\\
        \hline
        \hline
        \multirow{3}{0.9cm}{\centering Llama-3 70B} & ToT & 31.3 & 52.1 & 24 \\
        \cline{2-5}
        & GoT & 32.7 & 53.6 & 26  \\
        \cline{2-5}
        & MDToC & \textbf{36.1} & \textbf{57.3} & \textbf{29} \\
        \hline
    \end{tabular}
    \caption{Accuracy achieved by MDToC, ToT, and GoT on the CHAMP, MATH, and Game‑24 datasets using eight language models: GPT‑3.5‑Turbo, GPT‑4o‑mini, GPT‑4‑Turbo, GPT‑4o, Mistral‑7B, Mistral‑8×22B, Llama‑3‑8B, and Llama‑3‑70B}
    \label{benchmark_evaluation}
\end{table}
To better understand the performance of our MDToC, Table \ref{benchmark_evaluation} shows the performance of our MDToC when using the same GPT models and other open-source LLMs across all three components. Across a broad sweep of language models, our MDToC consistently outperforms both ToT and GoT. With GPT‑4‑Turbo, accuracy rises from 52.7\% (ToT) and 54.2\% (GoT) to 59.0\% on CHAMP, from 80.4\% and 80.8\% to 86.9\% on MATH, and from 79\% and 81\% to 85\% on Game‑24. Notably, the gains are even larger for smaller models: GPT‑4o‑mini sees a 9.3‑point jump on CHAMP and an 8.9‑point boost on Game‑24, underscoring MDToC’s ability to compensate for limited parameter capacity.

The trend extends to open‑source LLMs. When applied to Mistral-7B, our MDToC improved MATH accuracy from 23.5\% to 27.2\% (+3.7\% increase), CHAMP accuracy from 18.1\% to 19.7\%, and Game-of-24 accuracy from 8\% to 9\%. For the stronger Mistral 8×22B model, MDToC yielded improvements on MATH (52.7\% to 55.4\%), CHAMP (31.9\% to 34.5\%), and Game-of-24 (22\% to 26\%). The Llama-3 family showed similar benefits: the 8B variant experienced increases on MATH (26.8\% to 30.6\%) and CHAMP (19.2\% to 21.5\%), while the 70B variant saw substantial gains on MATH (52.1\% to 57.3\%), CHAMP (31.3\% to 36.1\%), and Game-of-24 (24\% to 29\%). These consistent performance improvements of 2-5\% across four open-source LLMs demonstrate that our MDToC enhances community models, mirroring the benefits previously observed with GPT architectures.

\subsection{Problem types on MATH dataset}
\label{sec:geo_appendix}
\begin{figure}[htbp]
    \centering
    \resizebox{0.49\textwidth}{!}{%
    \begin{tikzpicture}
        \begin{polaraxis}[
            grid=both,
            minor grid style={dotted},
            major grid style={dashed},
            xtick={0,51.43,102.86,154.29,205.71,257.14,308.57},
            xticklabels={
                algebra,
                counting\_and\_probability,
                geometry,
                intermediate\_algebra,
                number\_theory,
                prealgebra,
                precalculus
            },
            ytick={20,40,60,80,100},
            legend pos=north east,
            legend style={xshift=50pt} 
        ]
            \addplot[blue, thick, mark=square*] coordinates {
                (0,      93.3)
                (51.43,  92.8)
                (102.86, 72.4)
                (154.29, 88.7)
                (205.71, 92.9)
                (257.14, 87.2)
                (308.57, 78.6)
                (0,      93.3) 
            };
            \addlegendentry{MDToC}
            
            \addplot[red, thick, mark=triangle*] coordinates {
                (0,      87.6)
                (51.43,  86.1)
                (102.86, 70.2)
                (154.29, 82.9)
                (205.71, 81.5)
                (257.14, 78.9)
                (308.57, 75.8)
                (0,      87.6) 
            };
            \addlegendentry{ToT}
        \end{polaraxis}
    \end{tikzpicture}
    }
    \caption{\textbf{Radar chart comparing our MDToC and ToT performance on the MATH dataset} in terms of the percentage accuracy—both evaluated with GPT-4-Turbo—on various math problems of this dataset. }
    \label{math_radar}
\end{figure}
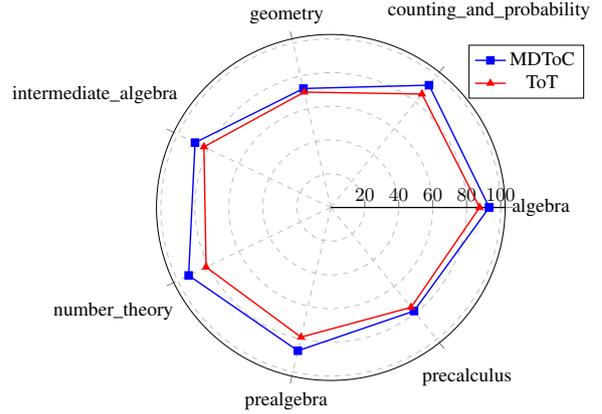
Figure \ref{math_radar} employs GPT-4-Turbo—selected for its superior performance over GPT-3.5-Turbo and GPT-4o-mini—to evaluate two prompting methods, ToT and our MDToC. Results were collected across seven math topics (algebra, counting and probability, geometry, intermediate algebra, number theory, prealgebra, and precalculus). MDToC outperformed ToT in five categories, showing notable leads in algebra (93.3\% vs. 87.6\%), intermediate algebra (88.7\% vs. 82.9\%), and counting and probability (92.8\% vs. 86.1\%). The methods performed similarly in geometry and pre-calculus (72.4\% vs. 70.2\% in geometry).

These experimental results show that while ToT and MDToC perform similarly on geometry-related and visual-understanding problems, notable differences emerge when more precise calculation steps are required. Geometry questions often hinge on spatial reasoning and visual understanding, domains in which both prompting methods perform equally well. In these tasks, the abstract thought evaluations encompassed by ToT appear sufficient to address the reasoning needed for shapes, angles, and other geometric relationships, while MDToC’s exclusive focus on calculations does not confer a distinct advantage. However, for algebra problems demanding intensive numeric manipulation, MDToC strongly outperforms ToT. This result aligns with MDToC’s design, which specifically targets explicit calculation steps to enhance accuracy in computation-heavy contexts.

\subsection{Problem types on CHAMP dataset}
\begin{figure}[htbp]
    \centering
    \resizebox{0.49\textwidth}{!}{%
    \begin{tikzpicture}
        \begin{polaraxis}[
            grid=both,
            minor grid style={dotted},
            major grid style={dashed},
            xtick={0,72,...,288},
            xticklabels={Combinatorics, Inequality, Number Theory, Polynomial, Sequence},
            ytick={20,40,60,80},
            width=10cm,
            legend pos=north east
        ]
            \addplot[blue, thick, mark=square*] coordinates {
                (0, 65.7)   
                (72, 60.8)  
                (144, 63.2) 
                (216, 48.2) 
                (288, 51.1) 
                (0, 65.7)   
            };
            \addlegendentry{MDToC}
            
            \addplot[red, thick, mark=triangle*] coordinates {
                (0, 55.7)   
                (72, 52.6)  
                (144, 56.5) 
                (216, 49.9) 
                (288, 49.1) 
                (0, 55.7)   
            };
            \addlegendentry{ToT}
        \end{polaraxis}
    \end{tikzpicture}
    }
    \caption{\textbf{Radar chart comparing our MDToC and ToT performance on the CHAMP dataset} in terms of the percentage accuracy—both evaluated with GPT-4-Turbo—across Combinatorics, Inequality, Number Theory, Polynomial, and Sequence.}
    \label{champ_radar}
\end{figure}
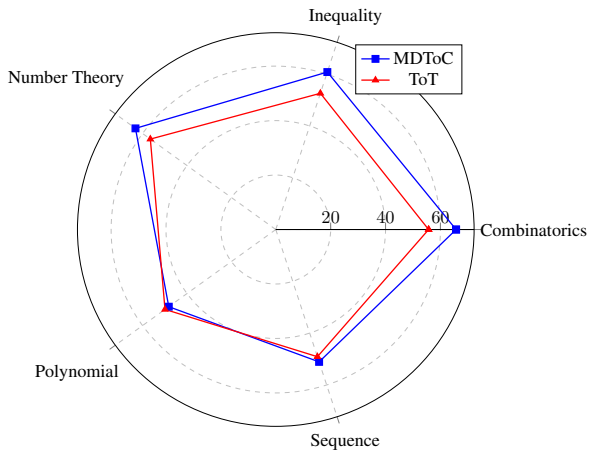

A comparative analysis of MDToC and ToT across five CHAMP dataset math topics in Figure \ref{champ_radar} shows MDToC's clear advantages in most categories. MDToC demonstrated significantly higher accuracy in Combinatorics (65.7\% vs. 55.7\%), Inequality (60.8\% vs. 52.6\%), and Number Theory (63.2\% vs. 56.5\%), highlighting its effectiveness in problems requiring detailed numeric calculations and methodical computation. While ToT showed a slight edge in Polynomial problems (49.9\% vs. 48.2\%), likely due to its strength in abstract symbolic manipulation, MDToC maintained superiority in Sequence problems (51.1\% vs. 49.1\%), suggesting that its explicit calculation framework better handles iterative, arithmetic-driven tasks. These results underscore how MDToC's focus on detailed numeric processes generally yields stronger performance in calculation-oriented mathematical problem-solving.
\end{document}